\documentclass[letterpaper, 10 pt, conference]{ieeeconf}  

\IEEEoverridecommandlockouts                              

\usepackage{graphicx} 
\usepackage[table]{xcolor}
\usepackage{pifont}
\usepackage{siunitx}
\usepackage{tikz}

\newif\ifshowieeedisclaimer
\showieeedisclaimertrue

\newcommand{\IEEEarxivnotice}{%
  \tikz[remember picture,overlay]{%
    \node[
      anchor=north,
      inner sep=2pt,
      outer sep=0pt,
      yshift=-0.3in,
      text width=0.98\textwidth,
      align=justify,
      draw=black,
      line width=0.4pt,
      fill=white,
      fill opacity=0.98
    ] at (current page.north) {%
      \footnotesize
      \setlength{\parskip}{0pt}%
      \setlength{\baselineskip}{9.5pt}%
      \textbf{Accepted for publication in IEEE ICDL 2026.}\ %
      \textcopyright\ 2026 IEEE. Personal use of this material is permitted. Permission from IEEE must be obtained for all other uses, in any current or future media, including reprinting/republishing this material for advertising or promotional purposes, creating new collective works, for resale or redistribution to servers or lists, or reuse of any copyrighted component of this work in other works.%
    };%
  }%
}

\title{\LARGE \bf
Estimating Central, Peripheral, and Temporal Visual Contributions to Human Decision Making in Atari Games
}

\author{Henrik Krauss$^{1}$ and Takehisa Yairi$^{2}$
\thanks{$^{1}$Henrik Krauss is with Department of Advanced Interdisciplinary Studies, The University of
Tokyo, Tokyo, 153-8904, Japan.
        {\tt\small henrik1.krauss@gmail.com}}%
\thanks{$^{2}$Takehisa Yairi is with the Research Center for Advanced Science and Technology, The University
of Tokyo, Tokyo, 153-8904, Japan.}
}

\begin{document}

\thispagestyle{empty}
\pagestyle{empty}
\maketitle
\ifshowieeedisclaimer
\IEEEarxivnotice
\fi
\thispagestyle{empty}
\pagestyle{empty}

\begin{abstract}
We study how different visual information sources contribute to human decision making in dynamic visual environments. Using Atari-HEAD, a large-scale Atari gameplay dataset with synchronized eye-tracking, we introduce a controlled ablation framework as a means to reverse-engineer the contribution of peripheral visual information, explicit gaze information in the form of gaze maps, and past-state information from human behavior. We train action-prediction networks under six settings that selectively include or exclude these information sources. Across 20 games, peripheral information shows by far the strongest contribution, with median prediction-accuracy drops in the range of 35.27--43.90\% when removed. Gaze information yields smaller drops of 2.11--2.76\%, while past-state information shows a broader range of 1.52--15.51\%, with the upper end likely more informative due to reduced peripheral-information leakage. To complement aggregate accuracies, we cluster states by true-action probabilities assigned by the different model configurations. This analysis identifies coarse behavioral regimes, including focus-dominated, periphery-dominated, and more contextual decision situations. These results suggest that human decision making in Atari depends strongly on information beyond the current focus of gaze, while the proposed framework provides a way to estimate such information-source contributions from behavior.
\end{abstract}

\section{Introduction}
\begin{figure}[t]
    \centering
    \includegraphics[width=\columnwidth, trim= 0 17mm 0 0]{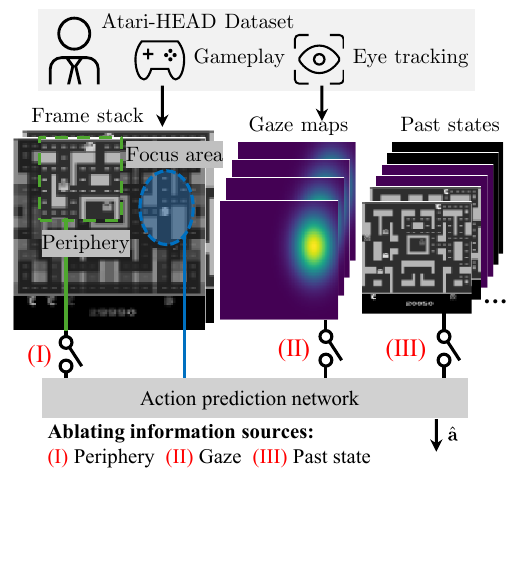}
    \caption{The controlled ablation framework performed in this study: Human gameplay and eye-tracking data from the Atari-HEAD dataset~\cite{zhang2020atari} are used to identify focus and periphery region, and construct gaze maps and past states. These information sources are included and excluded for the training of several action prediction networks to estimate their respective contributions.}
    \label{fig:intro}
\end{figure}
The frame problem~\cite{mccarthy1981some} highlights a core challenge in artificial intelligence of determining which aspects of a complex, changing environment are relevant for taking the next action. A closely related question in cognitive science is how humans allocate attention and integrate task-relevant information across space and time when making decisions in dynamic environments.

Recent work studies the relation of human behavior and visual attention to better understand human decision making. A systematic review by Brams et al.~\cite{brams2019relationship} found that experts across domains selectively allocating attention to task-relevant information. Ryu et al.~\cite{ryu2013role} showed that expert decision makers in basketball outperformed novices when predicting match situations under both central and peripheral viewing conditions. Jeong et al.~\cite{jeong2022difference} found that StarCraft experts exhibit broader attention and faster gaze movements than less skilled players, and DeCouto et al.~\cite{decouto2023role} showed that basketball players frequently update peripheral information with foveal vision while performance drops for larger eccentricities. Together, these studies suggest that human decision making depends on coordinated use of central and peripheral information rather than on gaze, i.e., the focus area alone.

Human attention and gaze information have also been used to improve intelligent agents. Zhang et al. identify learning from human attention as one general way of leveraging human guidance for deep reinforcement learning~\cite{zhang2019leveraging}. Practically, gaze information has been used to enhance human action estimation in autonomous driving~\cite{liu2019utilizing}, imitation learning (IL) for visual quadrotor navigation~\cite{bera2021gaze}, and visual robot manipulation from human demonstrations~\cite{chuang2025look}. In Atari, Attention-Guided Imitation Learning (AGIL) improved action prediction by augmenting IL with gaze information~\cite{zhang2018agil,zhang2020atari}, while Selective Eye-Gaze Augmentation (SEA) introduced a gating mechanism that learns when gaze should and should not be used~\cite{thammineni2023selective}. These studies show that gaze can be useful for training agents, but they mostly exploit gaze as a performance-enhancing signal rather than estimating the relative contribution of different human information sources.

Guo et al.~\cite{guo2021machine} compared machine and human attention and found that human attention remains robust in situations where agent attention can fail, but their interpretation is restricted to the focus area and therefore leaves covert, i.e., peripheral attention unresolved. Some previous studies come closer to the question of how focal and peripheral information contribute differently. Larson and Loschky~\cite{larson2009contributions} found that peripheral vision contributed more than central vision to maximal scene understanding from brief viewing, although central vision was more efficient when equal image area was compared. Nuthmann~\cite{nuthmann2014regions} showed in real-world object search that performance remained largely intact without foveal vision, whereas limiting access to extrafoveal information substantially impaired search. In our previous work, human attention maps derived from Atari gameplay showed high action-prediction performance even under high sparsity. These attention maps exhibited greater similarity to time-integrated human gaze than to artificial agent's attention, but clearly extend beyond the focus area~\cite{krauss2026revealing}. This points to the importance of studying covert attention, i.e. peripheral visual information, for human decision making as well.

However, despite these findings, we still lack a framework for estimating how different information sources contribute to human decisions. Existing gaze-augmented IL models often include gaze in a largely black-box way, and even approaches that learn to gate gaze do not directly explain when and why human experts rely on focal information, peripheral context, or short-term temporal context. In particular, the relative contributions of focus, periphery, and past-state information to human decision making remain insufficiently understood.

In this paper, we argue that these contributing factors can be estimated, or reverse-engineered, from human behavior with eye-tracking data. We use Atari gameplay data with synchronized gaze measurements to estimate central, peripheral, gaze, and temporal information. Estimating these factors is interesting for two main reasons: Firstly, to advance the understanding of human behavioral modes in decision-making. Secondly, to leverage this knowledge to build more intelligent artificial agents that can learn from human attention patterns during decision making beyond feeding gaze information into a black-box model.

The contributions of this paper are as follows:
\begin{enumerate}
\item We introduce a controlled ablation framework for analyzing peripheral, gaze, and past-state contributions to human action prediction in Atari. The approach is visualized in Fig.~\ref{fig:intro}.
\item We provide an analysis and discussion of how these information sources affect prediction performance over 20 Atari games.
\item We complement this with a cluster analysis of potential behavioral modes and an exploratory single-subject analysis.
\end{enumerate}

Section~\ref{sec:methods} describes the controlled ablation study, network architecture, dataset processing, training procedure, and cluster-analysis methodology. Section~\ref{sec:results} presents the action-prediction, cluster-analysis, and single-subject results. Section~\ref{sec:conclusions} concludes the paper with a summary and statement of limitations.

\section{Methods}
\label{sec:methods}
\subsection{Approach for a Controlled Ablation Study}
\label{subsec:approach}
We estimate central, peripheral, and visual contributions to human action prediction when playing Atari games by training IL models under different configurations. Particularly, a controlled ablation study is performed over the following three information sources:
\begin{enumerate}
\item \textbf{Periphery information:} This denotes visual content outside a gaze-centered region of a radius of ca. $6^\circ$ visual angle. We choose a relatively wide central region to make this a conservative test of peripheral-information effects.
Parafoveal vision is commonly described as extending to about $4$--$5^\circ$~\cite{nuthmann2014regions}, while scene-perception work has contrasted central and peripheral contributions using nearby radii in the $5^\circ$ to $7.4^\circ$ range~\cite{larson2009contributions}. When periphery information is disabled, pixels outside this region are replaced with the global mean frame.
\item \textbf{Gaze information:} Gaze maps precomputed from eye-tracking data are included as additional inputs. Specifically, we use four gaze maps per frame, each generated by convolving the measured gaze points after the last and up to the current frame with Gaussian distributions (standard deviation corresponding to $1^\circ$, $3^\circ$, $5^\circ$, and $10^\circ$ of visual angle, respectively) to provide gaze information at different spatial scales.
\item \textbf{Past-state information:} Visual and gaze inputs from three earlier sampled states together with the current state, using a stride of 15 frames, corresponding to approximately the previous $2.25$\,s of gameplay. This captures the time window of ~2–3 seconds for which humans report low difficulty of integrating visual information, as reported by Fairhall et al.~\cite{fairhall2014temporal}.
\end{enumerate}

In total we study six configurations of including or excluding the information sources stated above. The game frames, gaze maps and periphery-removed frames are also visualized in Fig.~\ref{fig:architecture}. The resulting six configurations are listed in Table~\ref{tab:model_configs} and denote all possible combinations, except those where periphery is excluded but gaze information not appended. These two configurations are not considered as interpreting their performance may be unclear, because the image with masked periphery already implicitly encodes gaze location.

\definecolor{modelA}{HTML}{4192D9}
\definecolor{modelB}{HTML}{004B8D}
\definecolor{modelC}{HTML}{7ED957}
\definecolor{modelD}{HTML}{137547}
\definecolor{modelE}{HTML}{FFBE00}
\definecolor{modelF}{HTML}{FD7400}

\begin{table}[t]
\centering
\caption{Model configurations of including and excluding different information sources as used in the controlled ablation study.}
\label{tab:model_configs}
\renewcommand{\arraystretch}{1.0}
\setlength{\tabcolsep}{3pt}
\begin{tabular}{c c c c p{3.0cm}}
\hline
Model & 1) Periphery & 2) Gaze & 3) Past state & Note \\
\hline
\cellcolor{modelA!35}\textbf{A} & \ding{109} & \ding{109} & \ding{109} & All information \\
\cellcolor{modelB!35}\textbf{B} & \ding{109} & \ding{109} & \ding{55} & comparable to AGIL~\cite{zhang2018agil} \\
\cellcolor{modelC!35}\textbf{C} & \ding{109} & \ding{55} & \ding{109} & \\
\cellcolor{modelD!35}\textbf{D} & \ding{109} & \ding{55} & \ding{55} & Standard IL \\
\cellcolor{modelE!35}\textbf{E} & \ding{55} & \ding{109} & \ding{109} &  \\
\cellcolor{modelF!35}\textbf{F} & \ding{55} & \ding{109} & \ding{55} &  \\
\hline
\end{tabular}

\vspace{1mm}
{\footnotesize \ding{109} included, \ding{55} excluded.}
\end{table}

\subsection{Neural Network Architecture}
\label{subsec:architecture}
\begin{figure*}[t]
    \centering
    \includegraphics[width=\textwidth, trim=0 67mm 0 0]{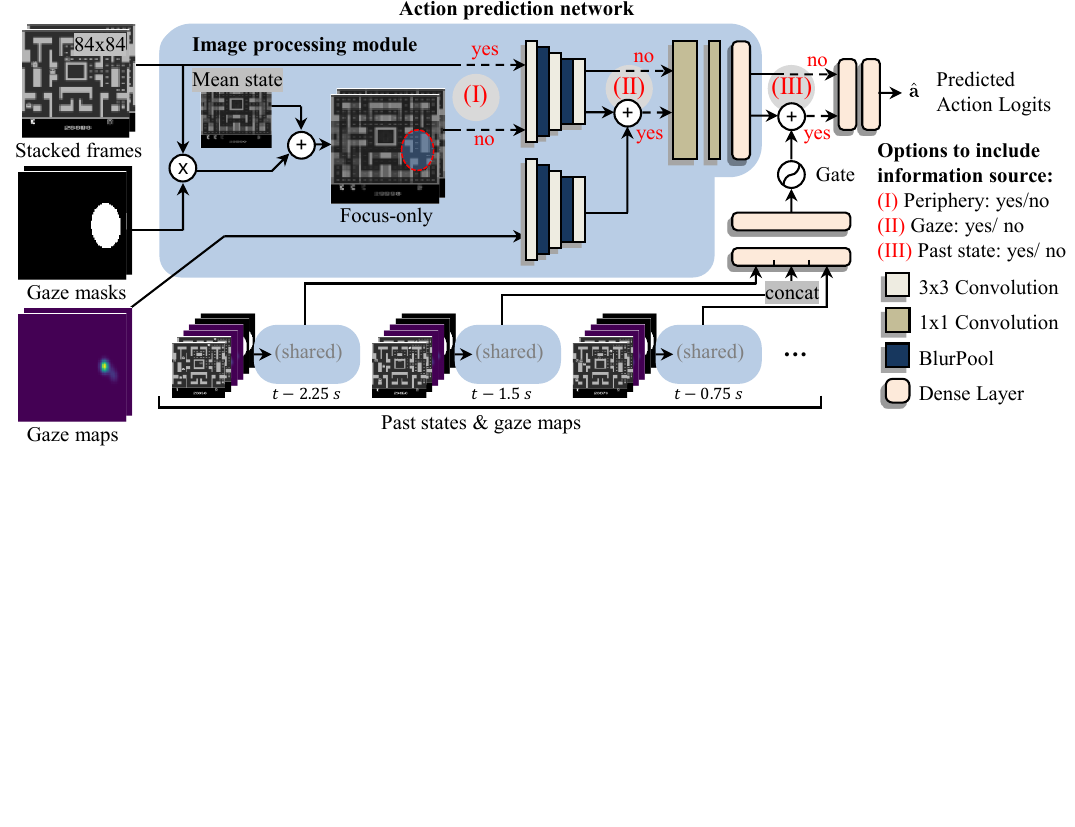}
    \caption{Architecture of the human action prediction network with three options of including (I) peripheral information, (II) gaze information, and (III) past-state information.}
    \label{fig:architecture}
\end{figure*}
The action prediction network architecture is shown in Fig.~\ref{fig:architecture}.
The architectural objective is twofold: Firstly, it enables inclusion and exclusion of the three information sources as defined in Sec.~\ref{subsec:approach}. Secondly, it 
tries to ensure that adding an informative source will increase action prediction accuracy. This is critical for our study, as we want to measure positive contributions of central, peripheral, and temporal visual information. This is non-trivial in deep learning, where additional inputs can sometimes degrade performance or generalization~\cite{huang2022modality}.

All variants share the same overall action-prediction structure. The current state input consists of two stacked, consecutive $84\times 84$ grayscale frames processed by a three-layer $3\times 3$ convolutional encoder (64, 64, and 32 channels) with anti-aliased downsampling (Blurpool~\cite{zhang2019making}, stride 2 on the first two layers), batch normalization, GELU activations, two $1\times 1$ convolution layers (32$\rightarrow$16$\rightarrow$8 channels), a 48-unit dense layer, followed by a linear action classifier. Variants with gaze augmentation add a parallel gaze-map encoder and fuse image and gaze features by averaging. Variants with past-state information encode the sampled earlier states with the same shared image processing module, compress their concatenation with dense layers, and combine them with the current representation through a learned sigmoid gate. This gate is a key design element, as it allows the network to ignore past-state information when it increases optimization difficulty without improving action prediction, so past-state information is used only to the extent that it is beneficial.

\subsection{Dataset Processing and Training}

We use Atari-HEAD~\cite{zhang2020atari}, a large-scale Atari human demonstration and eye-tracking dataset containing synchronized gameplay actions and gaze measurements across 20 Atari games. The dataset comprises approximately 117 hours of gameplay. We convert each game's record into an HDF5 replay buffer containing the current frame, next frame, human action, reward, terminal flag, source episode identifier, and variable-length gaze coordinates. Human actions are represented as one of 18 discrete Atari actions\footnote{All games use a fixed 18-way Atari joystick label space for comparability, though not all games exhibit every label.}, where combinations of button presses such as \texttt{UP+FIRE} are treated as separate action classes rather than decomposed into partial matches. Frames are resized to $84\times 84$ grayscale, and gaze maps are precomputed offline for efficient sampling.

Training uses two stacked frames per state. For variants with past-state information, each sample additionally contains a four-state temporal window with a stride of 15 frames. These past states are treated the same as the current state in terms of including periphery or gaze information, depending on the configuration. Valid temporal windows are restricted to remain within the same source recording so that stacked current/past inputs never cross session boundaries. All models are trained with batch size 64, cross-entropy loss, and AdamW optimizer at $10^{-3}$ initial learning rate, weight decay $10^{-2}$, and gradient clipping of 1.0. All models are trained for 150 quasi-epochs that consist of 200 randomly sampled batches over the training states. The learning rate is dropped to $10^{-4}$ after epoch 100. For model $m$, the training objective is
\begin{equation}
\mathcal{L}_{m} = - \frac{1}{|N|} \sum_{i \in N} \log p_{m}(a_i^{*}\mid x_i),
\end{equation}
where $N$ denotes the batch size, $a_i^{*}$ the demonstrated human action, and $x_i$ the input tuple including current frame stack, gaze maps, and past-state information. We use a block-based $90/10$ train/validation split with block size of 50 corresponding to at least $\SI{2.5}{s}$. Block-based split is preferred over fully random, as temporally close actions are strongly correlated. For each game, all six configurations A--F are trained jointly on one replay buffer (200 random batches per quasi-epoch, shared sampling, separate optimizers). Training and validation partitions contain states from all subjects for that game. Subject-specific models use per-subject buffers and are analyzed in Sec.~\ref{sec:results}. All non-terminal states with gaze data are used with identical hyperparameters. Joint training on one NVIDIA RTX A4000 GPU for 150 quasi-epochs per game took roughly \SI{3}{h}.

\subsection{Cluster-based Analysis of Model Predictions}

Beyond the analysis of the base performance levels of the six model configurations presented in Sec.~\ref{subsec:approach}, we perform a cluster analysis to capture patterns in the model's predictions across all games. For each state $i$, we form a six-dimensional feature vector
\begin{equation}
\mathbf{z}_i = \left[p_m(a_i^{*}\mid x_i)\right]_{m \in \{A,B,C,D,E,F\}},
\end{equation}
where $p_m(a_i^{*}\mid x_i)$ is the softmax probability assigned by model $m$ to the true human action. K-means with $k=5$ is fit on the pooled state set \emph{once across all games}. Clusters are then summarized through their mean per-model true-action probabilities and named retrospectively. Number of clusters $k=5$ is chosen empirically, based on the observation that for higher $k$, clusters emerged that exhibited highly similar patterns of true-action probabilities per models only at slightly different base probability levels. For the single-subject analysis, the same fitted clustering is reused to assign subject-specific states to the common cluster space. Cluster separability is summarized with silhouette scores on random subsamples, where higher values indicate better separation from neighboring clusters. For visualization, we compute a two-dimensional t-distributed stochastic neighbor embedding (t-SNE) on a random subsample of $10{,}000$ and $1{,}000$ pooled states for the overall chart and the individual games respectively, using perplexity of 80.

\section{Results}
\label{sec:results}
\subsection{Human Action Prediction Accuracies}
\begin{figure*}[t]
    \centering
    \includegraphics[width=\textwidth, trim=0 5mm 0 0]{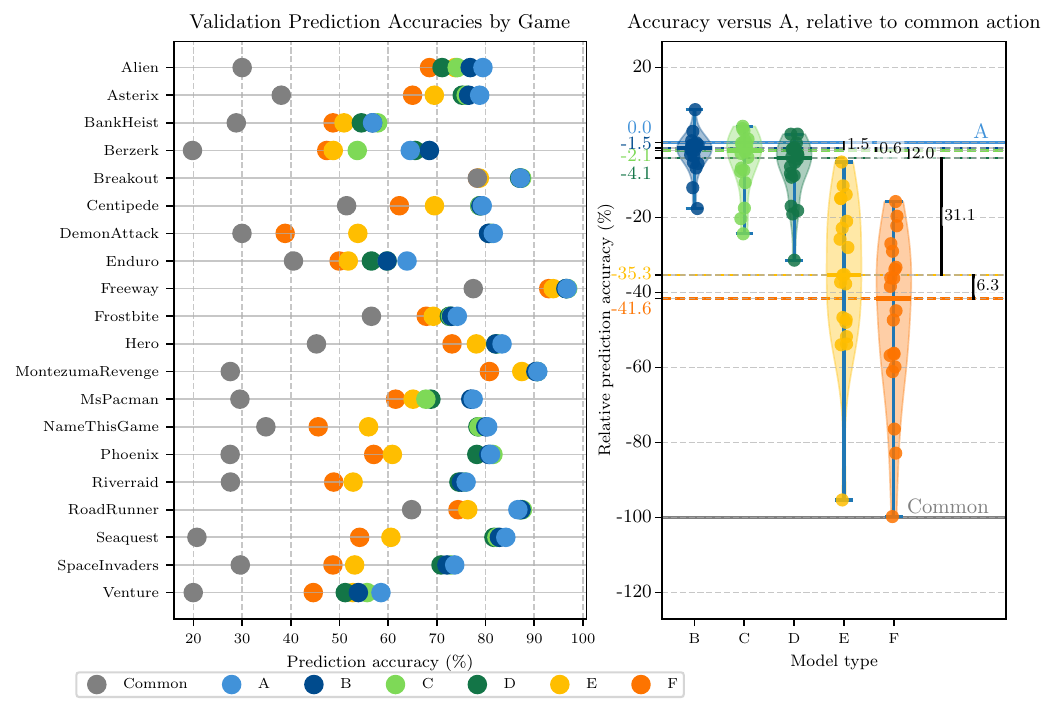}
    \caption{Validation action-prediction accuracies across games (left) and median relative performance drops with respect to model A, normalized by the A-common gap (right).}
    \label{fig:model_accuracies}
\end{figure*}

For game $g$ and model $m$, action prediction accuracy is defined as
\begin{equation}
\mathrm{Acc}_{g,m} = \frac{1}{|D_{g}^{\mathrm{val}}|}\sum_{i \in D_{g}^{\mathrm{val}}}
\mathbf{1}\!\left[\hat{a}_{i,m} = a_i^{*}\right],
\end{equation}
where $D_{g}^{\mathrm{val}}$ denotes the validation set size for game $g$, $a_i^{*}$ the demonstrated human action, and $\hat{a}_{i,m}=\arg\max_a p_m(a \mid x_i)$ the action predicted by model $m$. Accuracy therefore requires an exact match of the full discrete action class. Action distributions are imbalanced (e.g., frequent \texttt{NOOP}), but all six models beat a majority-class common-choice baseline on every game, so effects are not trivial majority prediction. Figure~\ref{fig:model_accuracies} summarizes per-game accuracies and median relative drops with respect to model A.

As stated in Sec.~\ref{subsec:architecture}, the architecture design aims that additional information sources yield better action prediction performance. To check this, we evaluate the following rule fulfillment across the 20 games, where $\mathrm{A}>\mathrm{B}$ means model A exhibits higher average prediction accuracy than model B:
\begin{center}
\begin{tabular}{ll}
$\mathrm{A}>\mathrm{B}$: 17/20 & $\mathrm{C}>\mathrm{D}$: 18/20 \\
$\mathrm{A}>\mathrm{C}$: 15/20 & $\mathrm{E}>\mathrm{F}$: 20/20 \\
$\mathrm{A}>\mathrm{E}$: 20/20 & $\mathrm{B}>\mathrm{D}$: 18/20 \\
$\mathrm{B}>\mathrm{F}$: 20/20 & $\mathrm{A}\text{--}\mathrm{F}>\text{common}$: 20/20 \\
\end{tabular}
\end{center}
We can see that these rules are overall well fulfilled, with mostly 0--3 outliers, except for 5 games where the addition of gaze information worsens prediction performance. This could be explained by model hyperparameters being the same for all games and not being adapted/optimized toward individual game complexities or varying dataset sizes. Overall, we consider these violations acceptable and include all trained models from all games in the numerical analysis.

\definecolor{dropPast}{HTML}{DCEBFA}
\definecolor{dropGaze}{HTML}{DDF4D0}
\definecolor{dropPeriph}{HTML}{FFE7BF}
\definecolor{diagGray}{HTML}{E6E6E6}

\begin{table}[h]
\centering
\caption{Median normalized pairwise accuracy differences (\%) across games. Normalized by the column model's accuracy difference with the common choice.}
\label{tab:accuracy_drops}
\renewcommand{\arraystretch}{1.08}
\setlength{\tabcolsep}{4pt}
\begin{tabular}{c c c c c c c}
\hline
 & A & B & C & D & E & F \\
\hline
A & \cellcolor{diagGray}+0.00 & +1.54 & +2.16 & +4.31 & +54.49 & +71.75 \\
B & \cellcolor{dropPast}-1.52 & \cellcolor{diagGray}+0.00 & -0.66 & +2.84 & +54.32 & +78.28 \\
C & \cellcolor{dropGaze}-2.11 & +0.66 & \cellcolor{diagGray}+0.00 & +1.98 & +41.52 & +64.55 \\
D & -4.13 & \cellcolor{dropGaze}-2.76 & \cellcolor{dropPast}-1.94 & \cellcolor{diagGray}+0.00 & +46.86 & +68.34 \\
E & \cellcolor{dropPeriph}-35.27 & -35.19 & -28.97 & -31.84 & \cellcolor{diagGray}+0.00 & +18.36 \\
F & -41.59 & \cellcolor{dropPeriph}-43.90 & -38.45 & -40.58 & \cellcolor{dropPast}-15.51 & \cellcolor{diagGray}+0.00 \\
\hline
\end{tabular}
\vspace{1mm}
\\
{\footnotesize \colorbox{dropPast}{past-state removal} \quad \colorbox{dropGaze}{gaze removal} \quad \colorbox{dropPeriph}{peripheral removal}}
\end{table}

To compare models beyond the A-referenced view in Fig.~\ref{fig:model_accuracies}, Table~\ref{tab:accuracy_drops} reports the full pairwise matrix of median normalized differences, where the normalization uses the gap between the column model and the most common action. The first column corresponds directly to the right panel of Fig.~\ref{fig:model_accuracies}, i.e. the median normalized drop relative to model $\mathrm{A}$. Removing past-state information leads to a median drop of 1.52\% ($\mathrm{A}\rightarrow\mathrm{B}$), removing explicit gaze information to 2.11\% ($\mathrm{A}\rightarrow\mathrm{C}$), removing both to 4.13\% ($\mathrm{A}\rightarrow\mathrm{D}$), removing peripheral information to 35.27\% ($\mathrm{A}\rightarrow\mathrm{E}$), and removing both, periphery and past-state information, drops by 41.59\% ($\mathrm{A}\rightarrow\mathrm{F}$). As visible in the right panel of Fig.~\ref{fig:model_accuracies}, these relative accuracy differences vary substantially across games, so the reported medians should not be understood as game-invariant effect sizes. The strongest decrease is nevertheless clearly associated with peripheral information removal, which is consistent with our previous findings on the importance of peripheral attention for human decision prediction~\cite{krauss2026revealing}. By contrast, adding explicit gaze information improves prediction performance only modestly in the present setting, despite earlier gaze-augmented IL work reporting substantial benefits~\cite{zhang2018agil,thammineni2023selective}.

The highlighted differences in Table~\ref{tab:accuracy_drops} that correspond to a single information source removal span 1.52--15.51\% for past-state removal, 2.11--2.76\% for gaze removal, and 35.27--43.90\% for peripheral removal. We regard the magnitudes of these ranges as a main finding, but the exact percentages should be interpreted with caution, since they are expected to differ slightly with architecture choices and do not represent fully clean decompositions. A main leakage issue is that information a human integrated over time may still remain inferable from the current peripheral scene, which would increase the apparent contribution of peripheral information while decreasing the apparent contribution of past-state information. This makes the larger $\mathrm{E}\rightarrow\mathrm{F}$ drop (15.51\%) particularly informative, since in that comparison peripheral information is already removed. Thus, the effective contribution of past-state information is likely closer to the upper end of the reported range than to the smaller drops observed for $\mathrm{A}\rightarrow\mathrm{B}$ and $\mathrm{C}\rightarrow\mathrm{D}$. Past true actions were also not included as model inputs explicitly to avoid auto-regressive fitting of the past actions. Although it is not clear how high this information leakage for autoregressive action fitting would be, some action history may still remain inferable from past visual states.

\subsection{Cluster Analysis for Human Action Prediction}
\begin{figure}[t]
    \centering
    \includegraphics[width=\columnwidth, trim= 0 5mm 0 0]{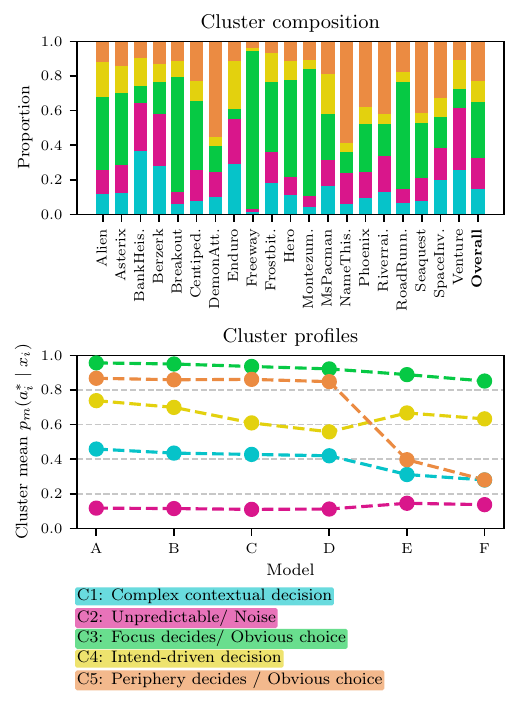}
    \caption{Cluster composition across games (upper), mean true-action-probability profiles (middle), and namings (lower) for the five clusters.}
    \label{fig:cluster_graphs}
\end{figure}
To analyze general human decision-making patterns with respect to the use of focal, peripheral, gaze, and past-state information, we cluster states in the shared six-dimensional model-response space introduced in Sec.~\ref{subsec:approach}, where each dimension is the true-action probability assigned by one model configuration. Figure~\ref{fig:cluster_graphs} shows the resulting cluster proportions across games and the mean six-model response profiles used for retrospective interpretation. The five cluster proportions are 14.9\%, 17.5\%, 32.5\%, 12.1\%, and 23.0\% overall, but distributions vary strongly between the different games.

Based on their mean true-action probability profiles, we name the five clusters as subjective heuristic labels (especially tentative for C1 and C4, silhouettes 0.10 and 0.08).
\textbf{C3 (Focus decides/ Obvious choice)} and \textbf{C5 (Periphery decides/ Obvious choice)} represent highly predictable modes. C3 exhibits the highest overall accuracy (up to $\sim$95\%) and minimal loss without peripheral data, suggesting foveal-driven, reactive play. In contrast, C5 shows a high base accuracy of $\sim$85\% that drops below 40\% when peripheral context is removed, identifying actions that are only obvious when the global game state is known.
\textbf{C1 (Complex contextual decision)} and \textbf{C4 (Intent-driven decision)} capture more nuanced strategies. C1 maintains a lower base accuracy ($\sim$40\%) with a reliance on peripheral context. C4 shows a unique profile where accuracy drops when gaze info is removed but increases when peripheral information is removed, suggesting that foveal focus and internal intent are more critical here than the overall game state. Finally, \textbf{C2 (Unpredictable/ Noise)} lies at a $\sim$10\% accuracy, possibly representing momentary unattentiveness or motor jitter.

\begin{figure}[t]
    \centering
    \includegraphics[width=\columnwidth,trim=0 5mm 0 0]{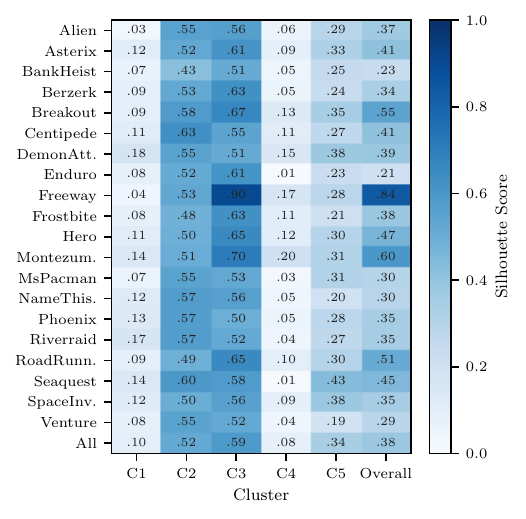}
    \caption{Mean silhouette scores per game and cluster, with an additional overall column.}
    \label{fig:cluster_silhouette}
\end{figure}
Figure~\ref{fig:cluster_silhouette} summarizes cluster separability for the same representation. The sampled overall silhouette score is 0.38, which suggests meaningful but not perfectly discrete separation. In particular, some clusters appear relatively compact and well separated (mean silhouette of C2 being 0.52 and C3 being 0.59), whereas others are much more diffuse (C1: 0.10 and C4: 0.08).
The silhouette heatmap in Fig.~\ref{fig:cluster_silhouette} shows that cluster separability varies substantially across games.

\begin{figure}[t]
    \centering
    \includegraphics[width=\columnwidth,trim=0 5mm 0 0]{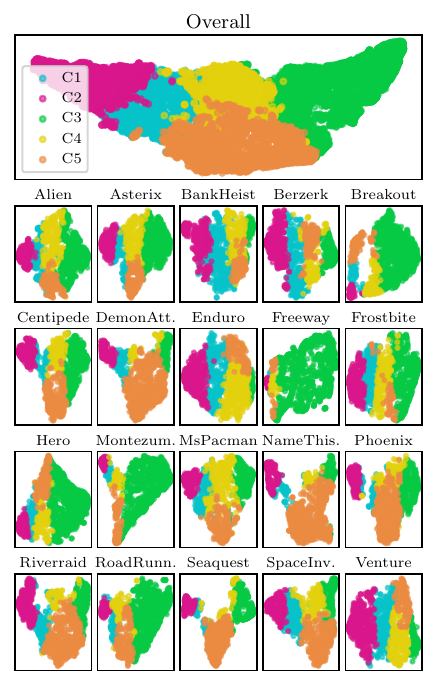}
    \caption{t-SNE visualization of the clustered six-model response space for all games (top) and per-game subsamples (bottom).}
    \label{fig:cluster_tsne}
\end{figure}
The t-SNE visualization in Fig.~\ref{fig:cluster_tsne} supports the same observation. The overall clusters occupy distinct regions but do not resemble separate islands, except from C3 that shows some separation from C4 and C5. For the individual games' t-SNE chart, we see some separated clusters, such as C2 for Centipede or Phoenix, as well as C3 for Seaquest. We therefore interpret the clusters as coarse behavioral regimes that help analyze patterns of decision making, but do not necessarily represent clear behavioral modes. For the more separated cluster regions in individual games, it is not entirely clear how much these correspond to behavioral modes or are a result of the specific game's mechanic.

\begin{figure*}[t]
    \centering
    \includegraphics[width=0.95\textwidth, trim= 0 5mm 0 0]{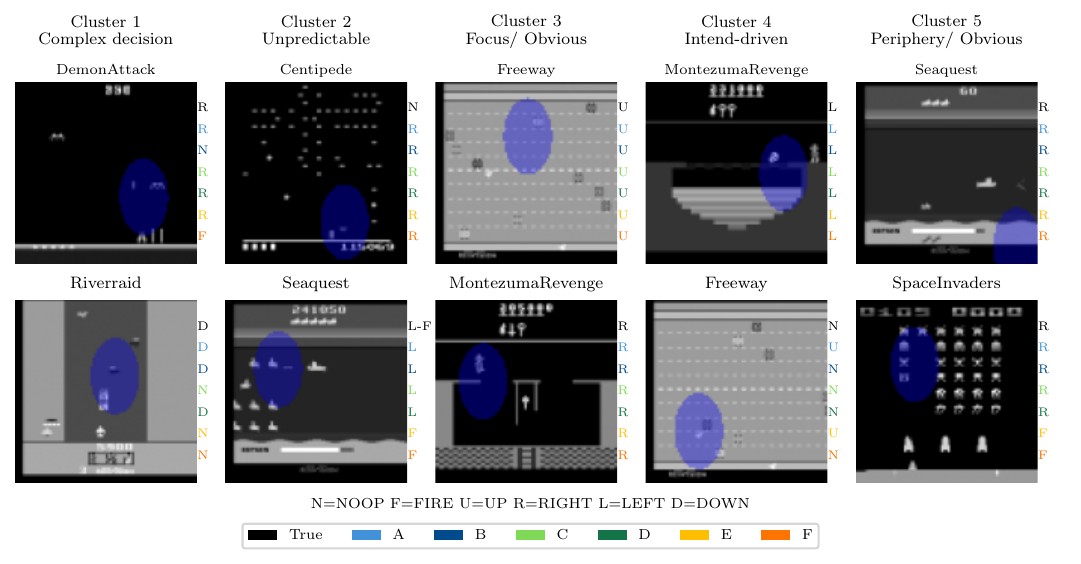}
    \caption{Example states from the two games with the highest silhouette score for each cluster. Focus region of six visual degrees is shaded in blue.}
    \label{fig:cluster_examples}
\end{figure*}

Fig.~\ref{fig:cluster_examples} shows two example scenes for each cluster from the two games with the highest silhouette scores in this cluster. We highlight three examples: (1) For the complex decision cluster C1, in Riverraid we observe the player's plane approaching a refueling station and a helicopter approaching flying direction. This indeed requires more complex decision making and the player decides to slow down, but several options are possible and some models predict no button press. (2) For the intent-driven cluster 4 in Freeway, we observe the human is looking at its chicken character being blocked from an incoming car and decides to press no button and wait. It is reasonable that the focus region here was most important here. (3) For the periphery-driven/ obvious choice cluster 5, we observe an expected pattern in SpaceInvaders. The player looks at the alien spaceships above while its character is located under a shield which is obvious from periphery or short-term memory. To attack, the player decides to go right, which is correctly predicted by all models except E--F, which predict firing immediately as they have limited or no information the player is currently below a shield.

\subsection{Single-Subject Analysis}

\begin{figure}[t]
    \centering
    \includegraphics[width=\columnwidth, trim = 0 5mm 0 0]{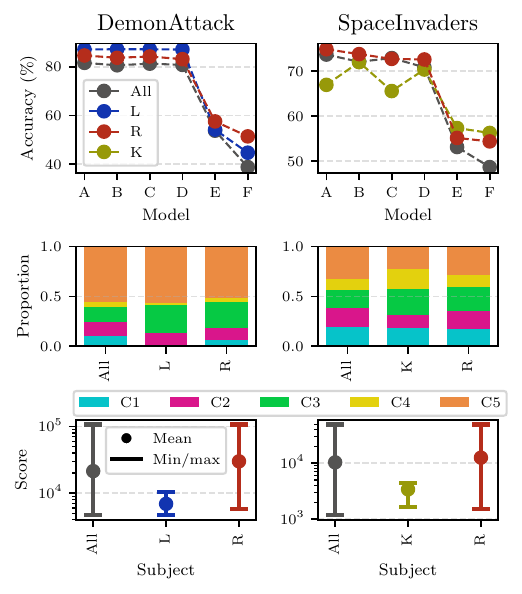}
    \caption{Single-subject comparison for DemonAttack and SpaceInvaders. Action-prediction accuracy across model types (top). Cluster composition (middle). Score statistics (bottom).}
    \label{fig:single_subject}
\end{figure}

As a limited follow-up analysis, we trained subject-specific models for two games with gameplay from at least two subjects and compared them with the corresponding all-subject models in Fig.~\ref{fig:single_subject}. In addition to the two single-subject models shown, gameplay from subject J is included in the all-subject model but was too limited to train an individual model. In DemonAttack, the high-performing subject R's actions are slightly less predictable than subject L's, except when periphery is removed. This could hint at subject R deciding more based on focus region, or could denote minor model performance differences due to different dataset sizes. The cluster composition differs only moderately from the all-subject model. This is similar to the results of SpaceInvaders, where subject R again performs strongly while the cluster composition remains broadly similar to the pooled model. Subject K in SpaceInvaders shows an unusual accuracy dropoff for model A and C, which may reflect overfitting of these higher parameter models rather than a meaningful behavioral difference.

Overall, these single-subject results do not support strong conclusions about subject-specific decision strategies. The dataset may be less suitable for analyzing models trained on individual subjects' data, as Atari-HEAD contains recordings from only four subjects overall, and individual games include data from only one to three of them. We therefore view the single-subject analysis in this study as exploratory only and do not interpret the observed differences as subject-level behavioral effects.

\section{Conclusions}
\label{sec:conclusions}

Understanding which information sources humans use to take action in dynamic visual environments is important both for explaining human decision making and for building more intelligent agents that learn from human behavior. In this paper, we addressed this question through a controlled ablation framework for human action prediction in Atari, using behavior with synchronized gaze data to reverse-engineer the contribution of peripheral visual information, explicit gaze information, and past-state information. Across 20 games, peripheral information showed by far the strongest effect, with median prediction-accuracy drops in the range of 35.27--43.90\% when removed. By contrast, gaze information removal yielded drops in the narrower range of 2.11--2.76\%. Past-state removal showed the broadest range, from 1.52\% to 15.51\%, where the upper end is likely more informative because it is measured after peripheral information has already been removed. At the same time, these drops vary substantially across games, so the reported ranges should not be understood as fixed effect sizes. A cluster analysis further suggests that these information sources are not used uniformly across states but depend on gameplay situation. Some clusters correspond to highly predictable focus- or periphery-dominated situations, whereas others capture more complex, intent-driven, or noisy decision regimes. Together, the study contributes a quantitative finding on peripheral-information dominance and a framework for estimating information-source contributions from behavior, suggesting that fast dynamic decisions rely heavily on extrafoveal context beyond gaze~\cite{nuthmann2014regions,decouto2023role}.

Several limitations should be considered when interpreting these findings. The single-subject analysis remained of limited value due to the small number of subjects in Atari-HEAD. The ablation design also cannot fully disentangle information leakage between sources. Information integrated over time may remain inferable from the current peripheral scene, and some action history may still be recoverable from past visual states even though past true actions are not provided explicitly. The present model set does not include a periphery-only condition that masks only the focus area while preserving the periphery. Adding such a configuration could offer a symmetric complement to peripheral removal in models E and F in an extended analysis.

Future work includes this periphery-only ablation, richer subject-level analysis, stricter temporal ablations and slow- versus fast-game leakage comparisons (with $\mathrm{E}\rightarrow\mathrm{F}$ as a partial bound), stronger inferential treatment of game-to-game variability, balanced metrics, and more action-specific cluster interpretation.

\bibliographystyle{IEEEtran}
\bibliography{references}

\end{document}